\documentclass[10pt,twocolumn,letterpaper]{article}

\usepackage[pagenumbers]{cvpr} % To force page numbers, e.g. for an arXiv version
\usepackage[accsupp]{axessibility} % Improves PDF readability for those with visual impairments.

\pdfoutput=1 
\usepackage{graphicx}
\usepackage{glossaries}
\usepackage[verbose]{placeins}
\usepackage{url}
\usepackage[hyphenbreaks]{breakurl}

\usepackage {algpseudocode}
\usepackage{algorithmicx}
\usepackage{algcompatible}
\usepackage{float}
\usepackage{multirow}

\newglossaryentry{SD}{
name={SDv2.1~\cite{rombach2022high}},
description={Stable Diffusion v2.1}}
\newglossaryentry{SD_nocite}{
name={SDv2.1},
description={Stable Diffusion v2.1}}
\newglossaryentry{Interven}{
name={Interven~\cite{bansal2022well}},
description={Ethical Interventions}}
\newglossaryentry{FairDiff}{
name={FairDiff~\cite{friedrich2023fair}},
description=FairDiffusion}
\newglossaryentry{BaseRAG}{
name={Base RAG},
description=Baseline RAG}
\newglossaryentry{TextAug}{
name=TextAug,
description={Textual Augmentation}}

\newglossaryentry{FRAG}{name={FairRAG},
description={Fair RAG}}

\usepackage[dvipsnames]{xcolor}

\usepackage[skip=4pt]{caption}

\definecolor{cvprblue}{rgb}{0.21,0.49,0.74}
\usepackage[pagebackref,breaklinks,colorlinks,citecolor=cvprblue]{hyperref}

\def\paperID{2242} % *** Enter the Paper ID here
\def\confName{CVPR}
\def\confYear{2024}

\title{FairRAG: Fair Human Generation via Fair Retrieval Augmentation}

\author{~Robik Shrestha$^1$\thanks{Work done during an internship at AWS AI Labs.} \quad Yang Zou$^2$ \quad Qiuyu Chen$^2$ \quad Zhiheng Li$^2$ \quad Yusheng Xie$^2$ \quad Siqi Deng$^2$ \\ $^1$AWS AI Labs \quad $^2$Amazon AGI \\ {\tt\small robikshrestha@gmail.com \quad \{yanzo,qychen,lzhiheng,yushx,siqideng\}@amazon.com}
}

\begin{document}
\definecolor{green}{rgb}{0.0, 0.5, 0.0}
\definecolor{red}{rgb}{0.82, 0.1, 0.26}

\maketitle
\begin{abstract}
Existing text-to-image generative models reflect or even amplify societal biases ingrained in their training data. This is especially concerning for human image generation where models are biased against certain demographic groups. Existing attempts to rectify this issue are hindered by the inherent limitations of the pre-trained models and fail to substantially improve demographic diversity. In this work, we introduce Fair Retrieval Augmented Generation (\gls{FRAG}), a novel framework that conditions pre-trained generative models on reference images retrieved from an external image database to improve fairness in human generation. \gls{FRAG} enables conditioning through a lightweight linear module that projects reference images into the textual space. To enhance fairness, \gls{FRAG} applies simple-yet-effective debiasing strategies, providing images from diverse demographic groups during the generative process. Extensive experiments demonstrate that \gls{FRAG} outperforms existing methods in terms of demographic diversity, image-text alignment, and image fidelity while incurring minimal computational overhead during inference.
\end{abstract}    
\section{Introduction}
\label{sec:intro}

\begin{figure}[ht!]
\footnotesize
 \centering     \includegraphics[width=0.42\textwidth]{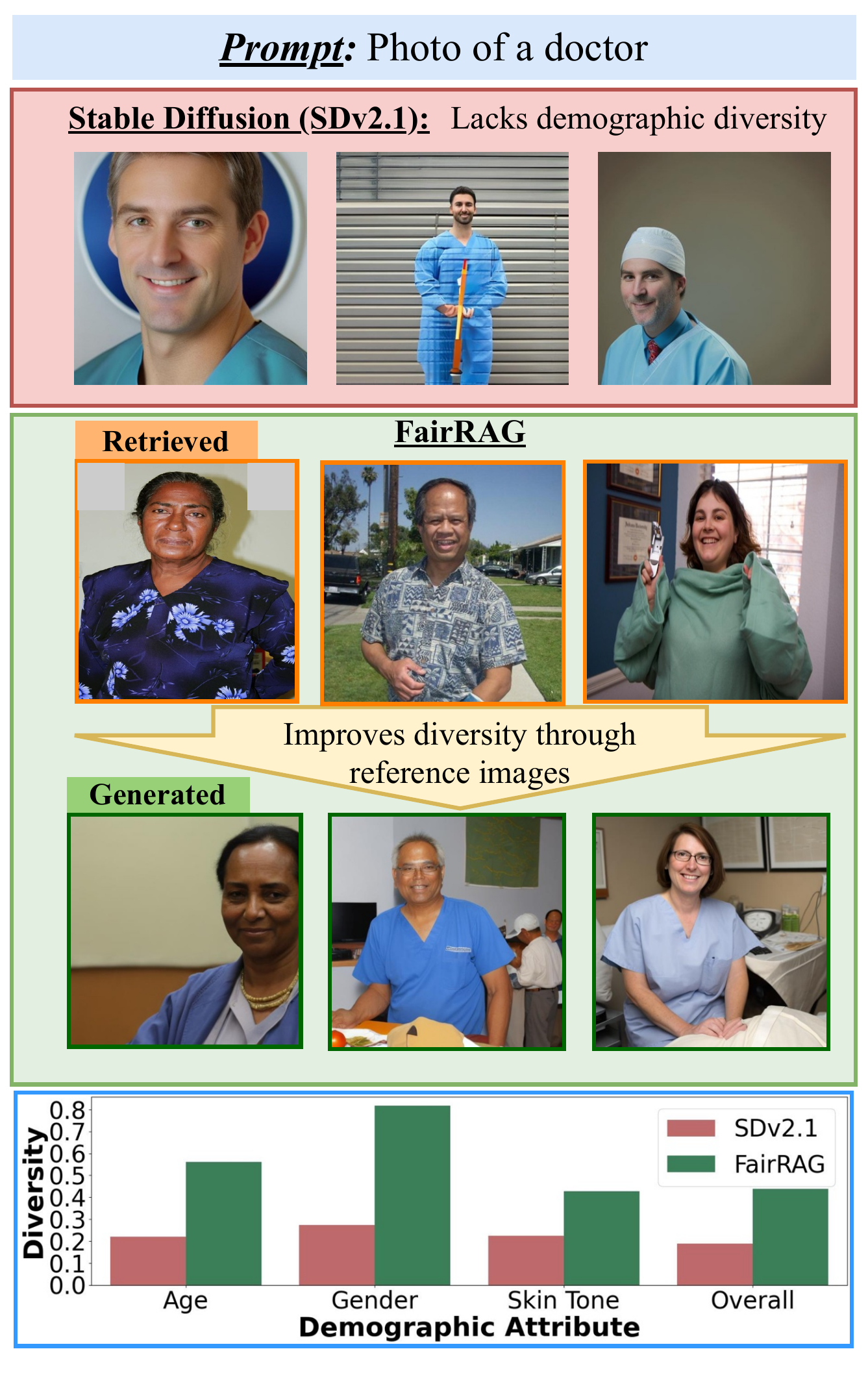}
  \caption{The proposed (\gls{FRAG}) framework improves demographic diversity (fairness in image generation) by conditioning generative models on external human reference images. As defined in Eq.~\ref{eq:diversity}, the diversity metric measures representation from different age, gender and skin tone groups.}
\label{fig:teaser}
\end{figure}

Generative artificial intelligence has witnessed rapid growth and adoption in a short span of time. In particular, diffusion-based text-to-image models are able to produce high-quality, photo-realistic images from textual prompts~\cite{ramesh2021zero,rombach2022high,saharia2022photorealistic,gafni2022make} and are thus increasingly being integrated into practical applications~\cite{shutterstockai, adobeAI}. However, this growing adoption also underscores the need to investigate and address fairness concerns. Specifically, text-to-image generation systems tend to mirror or even amplify societal biases in their training data, which is especially evident in human image generation~\cite{bansal2022well,luccioni2023stable,perera2023analyzing}. They exhibit biases against specific demographic groups in terms of age, gender and skin tone. For example, Stable Diffusion~\cite{rombach2022high} produces individuals with darker skin tones when prompted for workers from lower-paying occupations~\cite{bloomberg}. These tendencies result in adverse outcomes and diverge from the goals of equitable representation. There are some attempts to address this issue~\cite{friedrich2023fair,bansal2022well}, however, they do not adequately mitigate the dataset biases ingrained within the pre-trained models.

To address this, we introduce \textit{Fair Retrieval Augmented Generation} (\gls{FRAG}), which harnesses an external data source consisting of real human images from diverse age, gender, and skin tone groups to improve fairness. The \gls{FRAG} framework allows us to fix bias without the costly processes of fixing pre-training data or re-training backbone. Additionally, by expanding the external dataset, it can easily generalize to newer concepts, making it an extensible framework. \gls{FRAG} utilizes lightweight, yet effective mechanisms to improve fairness. First, \gls{FRAG} requires a way to condition the generative model on reference images. For this, we train a single linear layer that projects reference images into textual space to condition a frozen backbone. This circumvents the computational overhead in existing conditioning approaches, which either re-train the model~\cite{blattmann2022retrieval,chen2022re} or require test-time parameter tuning~\cite{gal2022image}. At inference, directly retrieving and conditioning on a set of images with the highest similarity score for a text prompt does not improve fairness because biases exist in the external database too. To address this, \gls{FRAG} consists of a fair retrieval system that utilizes efficient, post-hoc debiasing strategies to sample from diverse demographic groups. Compared to previous approaches~\cite{friedrich2023fair,bansal2022well} that fully rely on internal knowledge in the models, which can be biased, \gls{FRAG} is more steerable, explainable, and transparent in controlling demographic distributions for image generation.

We compare \gls{FRAG} against multiple methods in terms of the demographic diversity metric (\cf ~\cref{eq:diversity}), which assigns higher scores for fairer demographic representations. Compared to the best non-RAG method, \gls{FRAG} improves the diversity metric from 0.341 to 0.438. We also observe improvements in  image-text alignment (CLIP score~\cite{radford2021learning}): 0.144 to 0.146 and image fidelity (FID~\cite{heusel2017gans}): 74.1 to 51.8.

The contributions of this paper are as follows.
\begin{itemize}[noitemsep]
    \item We propose \gls{FRAG}, a novel framework to improve demographic diversity in human generation by leveraging reference images drawn from external sources.
    \item \gls{FRAG} employs lightweight conditioning and fairness-enhanced retrieval mechanisms that require minimal computational overhead.
    \item Experimental results show improvements over existing methods in terms of diversity, alignment and fidelity.
\end{itemize}
\section{Related Works} 
\noindent \textbf{Societal Biases in Diffusion Models.} \quad
Diffusion-based text-to-image generative models produce high fidelity, realistic images and have seen increasing adoption~\cite{ramesh2021zero,rombach2022high,saharia2022photorealistic,gafni2022make,shutterstockai, adobeAI}. However, they are trained on large-scale image-text datasets that contain harmful biases~\cite{schuhmann2022laion,bianchi2023easily}. Several works study how this causes the text-to-image generative systems to also be biased against specific demographic groups~\cite{perera2023analyzing,friedrich2023fair,bloomberg,bianchi2023easily}. Some recent works attempt to mitigate these issues, for instance, by editing the text prompt to encourage diversity~\cite{bansal2022well} or by guiding the generative process to balance out the representations from different groups~\cite{friedrich2023fair,gandikota2024unified}. However, such methods do not substantially mitigate the effects of the biased associations embedded within the models. To tackle this, \gls{FRAG} leverages external references that lessen such biases, \ie, contain samples from diverse groups to improve fairness in generation.

\noindent \textbf{Conditioning Text-to-Image Diffusion Models.} \quad
There are existing approaches to condition on visual references~\cite{gal2022image,ruiz2023dreambooth,zhang2023adding,xiao2023fastcomposer}. Some employ test-time tuning which is computationally expensive since it requires changing model parameters at inference~\cite{gal2022image,ruiz2023dreambooth,kumari2023multi}. Tuning-free methods avoid this by employing a separate adaptor module that is already trained for conditioning~\cite{zhang2023adding,mou2023t2i,xiao2023fastcomposer,shi2023instantbooth}. \gls{FRAG} is also a tuning-free method. However, compared to the heavier adaptor modules used in prior works, \gls{FRAG} uses a lightweight linear conditioning layer. Another concurrent work: ITI-Gen~\cite{zhang2023iti} learns prompt embeddings from visual references for conditioning. However, this entails learning a separate embedding per concept, which is not scalable. \gls{FRAG}, on the other hand, trains the conditioning module once and re-uses it to transfer demographic attributes and contextual information from new images at inference, making it more general.

\noindent \textbf{Retrieval Augmented Generation (RAG).} \quad
RAG-based methods retrieve relevant items from external sources to condition the generative process~\cite{lewis2020retrieval,thulke2021efficient,cai2022recent}. RAG has only recently been explored for diffusion-based models. For example, a recent RAG-based method~\cite{blattmann2022retrieval} shows
improvements in image quality and style transfer. Another work, Re-Imagen~\cite{chen2022re}, shows the efficacy of RAG in generating rare and unseen entities. Compared to these works, \gls{FRAG} is more suitable for fair human generation. Unlike previous approaches, \gls{FRAG} has a retrieval mechanism designed to improve demographic diversity and does not require costly retraining of backbone to support conditioning.

\section{Fair Retrieval Augmented Generation}

\begin{figure*}[ht!]
 \centering
        \includegraphics[width=0.9\linewidth]{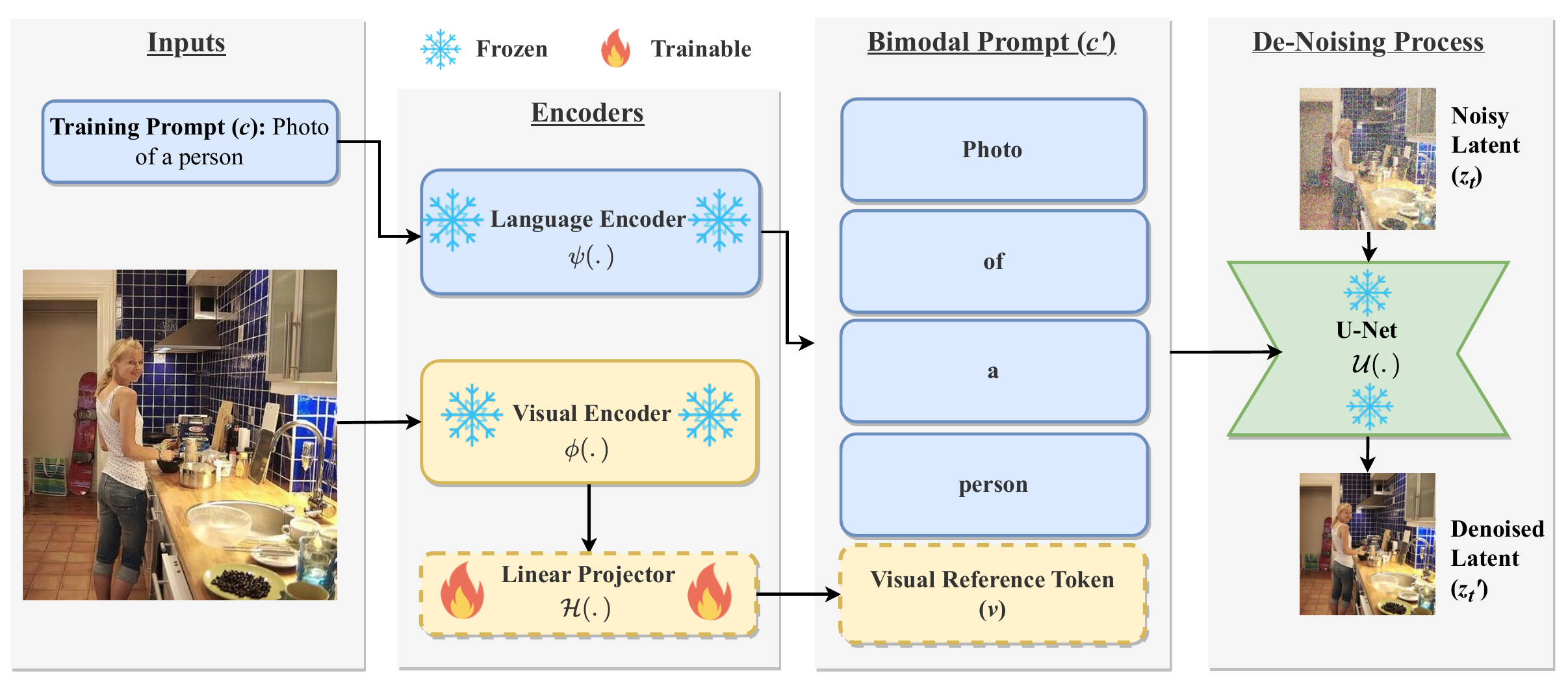}
        \caption{We train the linear projector $\mathcal{H}(.)$ using a denoising loss on the latent space while keeping the backbone model frozen. To train $\mathcal{H}(.)$, we sample images uniformly from each demographic group, pairing each image with the prompt: \textit{Photo of a person}.}
    \label{fig:training}
\end{figure*}
We propose the \gls{FRAG} framework to improve fairness in human image generation by using demographically diverse reference images. To achieve this, \gls{FRAG} requires a mechanism to condition the pre-trained backbone, which is enabled by training a lightweight linear encoder while keeping the backbone frozen (\cf \cref{sec:conditioning} and \cref{fig:training}). At inference, \gls{FRAG} uses simple, post-hoc debiasing strategies to improve fairness, including balanced sampling and query modification to for fair retrieval (Sec.~\ref{sec:fair_retrieval}) and a transfer instruction to enhance the generative process (Sec.~\ref{sec:image_generation}).

\subsection{Linear Conditioning Mechanism}
In this section, we first give the background of the backbone generative model, which is kept frozen for both training and inference. Then we describe our mechanism that conditions the frozen backbone on the references (Fig.~\ref{fig:training}).

\noindent \textbf{Frozen Backbone.} \quad
Our frozen backbone is a pre-trained text-to-image latent diffusion model---Stable Diffusion (SD)~\cite{rombach2022high}. It reverses noises applied to the latent embeddings of images. SD contains a variational autoencoder (VAE)~\cite{kingma2013auto}: $\mathcal{E}(.)$, a text encoder: $\Psi(.)$ and a U-Net~\cite{ronneberger2015u}:  $\mathcal{U}(.)$. Specifically, VAE encodes images $x$ to produce latent representations $z$. During the forward diffusion process, SD uses a noise scheduler to sample a timestep $t$, and applies Gaussian noise: $\epsilon_t \sim \mathcal{N}(0, 1)$ to $z$. During the backward diffusion process, $\mathcal{U}(.)$ estimates the noise ($\epsilon'_t$) added to the latent, enabling image generation via iterative denoising. The denoising process can also be conditioned on text prompt: $c$ encoded by the text encoder. Specifically, $c$ is fed alongside the noisy latent: $z_t$ into $\mathcal{U}(.)$ to control the denoising process. During inference, one can feed in random Gaussian noise and text prompt through the model to generate images.

\noindent \textbf{Conditioning Module.} \quad
\label{sec:conditioning}
As shown in Fig.~\ref{fig:training}, we use a linear projector: $\mathcal{H}(.)$ to condition the backbone on retrieved human references. $\mathcal{H}$ projects the reference image into a text-compatible token, augmenting the text prompt with additional information for conditioning. Let $x$ be the reference image, $v = \phi(x)$ be the visual embedding obtained from a CLIP image encoder~\cite{radford2021learning} and $c = (w_1, w_2, ..., w_n)$ be the text prompt encoded via a CLIP text encoder $\Psi(.)$. The linear projector: $\mathcal{H}(.)$ projects $v$ into a conditioning vector (token): $\mathcal{H}(v)$, which is concatenated with $c$ to obtain a bimodal prompt: $c' = (w_1, w_2, ..., w_n, \mathcal{H}(v))$. This retrieval-augmented text prompt is then fed into the U-Net to condition the denoising process.

\begin{figure*}[ht!]
\centering
\includegraphics[width=0.88\linewidth]{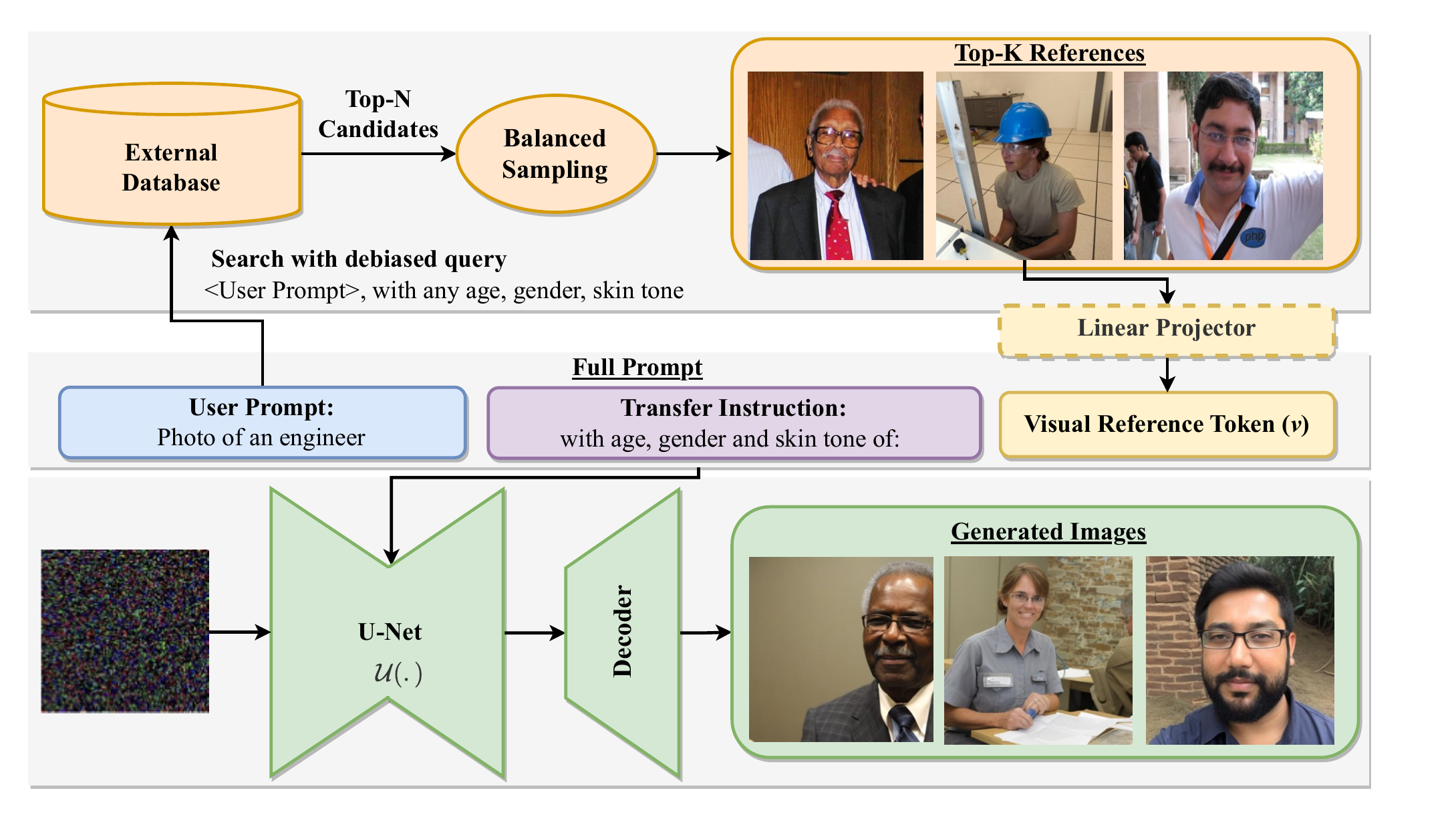}
\caption{During inference, \gls{FRAG} constructs a debiased query to retrieve Top-$N$ candidates for a given prompt. Using their demographic group annotations, \gls{FRAG} then selects a balanced set of $K$ images with high demographic diversity for conditioning. The full bimodal prompt consists of: a) the original user prompt, b) a transfer instruction and c) the projected visual reference token. This bimodal prompt is used within the cross-attention layers of the U-Net to condition the generative process.}
    \label{fig:inference}
\end{figure*}

\noindent \textbf{Training Procedure.} \quad
We train $\mathcal{H}(.)$ with the denoising loss in the latent space (Fig.~\ref{fig:training}). At timestep $t$, we train $\mathcal{H}(.)$ with the following denoising loss~\cite{rombach2022high}:

\begin{align}
~\mathcal{L} = \mathbb{E}_{z \sim \mathcal{E}(x), \epsilon \sim \mathcal{N}(0, 1),t } \left [ || \epsilon - \epsilon'(z_t, t, c') ||_2^2 \right ].
\end{align}
We pair the images with a simple text prompt: \textit{Photo of a person} during training, avoiding the usage of detailed captions that may not always be accessible.

\subsection{Fair Retrieval System}
\label{sec:fair_retrieval}
During inference, \gls{FRAG} ensures that the reference images are demographically diverse by using simple post-hoc debiasing techniques that do not require model re-training. The conventional approach of retrieving the Top-$K$ most similar images for a given query, as employed in prior RAG frameworks~\cite{chen2022re,blattmann2022retrieval}, does not ensure diversity. To address this limitation, \gls{FRAG} adopts a two-step process. First, it retrieves a larger set of $N$ candidate images ($N > K$), then, it performs balanced group sampling to obtain a balanced set of $K$ references to condition the model.

\noindent \textbf{Top-$N$ Retrieval with Debiased Query.} \quad
To obtain $N$ demographically diverse candidate images, \gls{FRAG} constructs a \textit{debiased query} by appending the original text prompt with the following phrase: \textit{with any age, gender, skin tone}. This simple query modification improves fairness in retrieval while maintaining consistency with the prompt. 

\noindent \textbf{Top-$K$ Selection via Balanced Sampling.} \quad While the debiased query improves diversity, the candidate images may not be ordered in a balanced manner. For this, \gls{FRAG} applies a \textit{balanced sampling strategy} and selects a set with $K$ images for conditioning. We store the prediction for each demographic attribute: $a \in $ \{\textit{age, gender, skin tone}\} for each image and use them for balanced sampling. For this, let us denote \textit{an intersectional group} as $g$, which is a tuple of the attribute values \eg, $g$ = \textit{(25 year, dark-skinned, male)} and \textit{an individual group} corresponding to the demographic attribute ($a$), as: $g[a]$. For instance, $g[age]$ = \textit{25 year} is an individual age group. Next, let $G$ be the set of unique intersectional groups in the Top-$N$ candidates and $m_{g[a]}$ be the number of times an individual group $g[a]$ appears in $G$. Then, the sampling weight for $g$ is:
\begin{align}
    w_g = \left [\sum_{a} \frac{m_{g[a]}}{n_a} \right ]^{-1},
\end{align} 
which is high if $g$ has individual groups that are rare and low if they are frequent. Here, $n_a$ is the total number of possible values for $a$, used for normalization (\eg, $n_{gender} = 2$). This strategy thus provides higher priority to the demographic groups that are underrepresented. 
 
\subsection{Image Generation}
\label{sec:image_generation}
As shown in Fig.~\ref{fig:inference}, \gls{FRAG} projects the reference images through $\mathcal{H}$ and adds a text instruction to enhance attribute transfer while generating images. Given an example prompt: \textit{Photo of an engineer}, \gls{FRAG} constructs an extended bimodal prompt: \textit{Photo of an engineer, with age, gender and skin tone of: $v$}, which contains the instruction: $\textit{with age, gender and skin tone of: }$ to improve the conditioning process. This method does not explicitly specify the exact age, gender or skin tone values, yet helps the model transfer those attributes from the reference images. This extended prompt is used within the cross-attention layers of $\mathcal{U}(.)$ to condition the model at each denoising step.

\section{Experiments}
\label{sec:experimental_setup}

\subsection{Experiment Setup}
We compare \gls{FRAG} against other baselines for the task of diverse human generation using neutral text prompts that do not specify any demographic groups, but still exhibit bias. We evaluate diversity among three \textit{demographic attributes}: age, gender and skin tone. We use a modified version of the \textit{demographic groups} presented in FairFace~\cite{karkkainen2019fairface} for: age (\textit{$<$ 20, 20-29, 30-39, 40-49, 50-59, 60+}) and gender (\textit{male} and \textit{female}). For skin tone analysis, we use the 10-point Monk Skin Tone (MST) scale~\cite{skintonegoogle}.

\noindent \textbf{Evaluation Prompts.} \quad The evaluation set of \gls{FRAG} consists of test prompts with professions that exhibit bias with respect to different demographic groups. These prompts are classified into 8 categories, including: 6 artists (\eg, a dancer), 6 food and beverage (F\&B) workers (\eg, a cook), 9 musicians (\eg, a guitarist), 6 security personnel (\eg, a guard), 9 sports players (\eg, a tennis player), 12 STEM professionals (\eg, an engineer), 7 workers (\eg, a laborer) and 25 from other professions (\eg, a politician). Please refer to ~\cref{appendix:eval_set} for the full list of prompts. For the main experiments, we use the template \textit{Photo of $<$profession$>$}, \eg, \textit{Photo of a doctor} to create the prompt.

\noindent \textbf{Demographic Diversity Metric.} \quad
We use normalized entropy as our diversity metric, reporting the intersectional and individual values, where higher values indicate more equitable representation across demographic groups. Intersectional diversity is computed over the unique combinations of age, gender and skin tone groups, and individual diversity score is computed separately per attribute, \eg, the gender diversity score is the normalized entropy for \textit{male} and \textit{female} categories. Specifically, let $p_i$ be the proportion of images generated for $i^{th}$ group, then, the diversity score is the entropy of the group memberships normalized by a uniform distribution, with a maximum value of 1:
\begin{align}
\mathcal{D} = \frac{\sum\limits_{i=1}^{n} p_i\log \left [p_i \right ] }{\frac{1}{n} \log(\frac{1}{n})},
\label{eq:diversity}
\end{align}
\noindent where, $n$ is the total number of possible groups. We compute the metrics over the images generated for each prompt, reporting the average across all prompts. 

\noindent We employ existing prediction systems for age, gender and skin tone classification. For age classification, DeepFace~\cite{serengil2021hyperextended} predicts an integer age value, which we map to our age-range categories. We follow Dall-Eval~\cite{cho2023dall} for gender and skin tone classification. Specifically, we use the CLIP model (ViT/L-14) with two classifier prompts: \textit{photo of a male or a man or a boy} and  \textit{photo of a female or a woman or a girl}, using the highest scoring prompt to determine the gender. For skin tone classification, we detect skin pixels within the facial region, and determine the Monk Skin Tone value that is closest to the average color of the skin region~\cite{kolkur2017human}. When computing the diversity metrics, we account for the cases where the methods fail to generate any human face. In such instances, we treat the images as belonging to the  most prevalent demographic group, thereby imposing a penalty on the diversity score.

While we follow previous research~\cite{friedrich2023fair, cho2023dall} for demographic group classification, we acknowledge their limitations as well. First, we perform skin tone analysis, but refrain from making inferences about the race. This choice is driven by the recognition that racial identity can be influenced by social and political factors~\cite{cho2023dall} and whether one can predict race from visual information alone is debatable. We leverage skin tone, a lower-level image feature, in an attempt to conduct more objective assessment. More precisely speaking, we consider the \textit{apparent} skin tone from RGB images, in the absence of access to true skin tone where more rigorous process needs to be applied (\eg, lab controlled data collection via spectrophotometers). Second, we employ a simplified binary gender classification, even though gender is known to encompass a broader spectrum~\cite{genderidentity}. This is because estimating gender from a wider range of possibilities based solely on appearance can potentially reinforce appearance-related stereotypes. While our studies and discussions of gender diversity in this work are limited to apparent binary genders, the framework we devise may be generalized to alternative definitions.

\noindent \textbf{Alignment and Fidelity Metrics.} \quad
We use CLIP score~\cite{radford2021learning}, \ie, the cosine similarity between the text embeddings and the image embeddings to compute image-text alignment. We report the Fréchet Inception Distance (FID)~\cite{heusel2017gans} between the generated images and real distribution, approximated by sampling a fixed number of real images per prompt.

\noindent \textbf{Training and Retrieval Sets.} \quad We use images containing humans from two datasets: MSCOCO~\cite{lin2014microsoft} and OpenImages-v6~\cite{openimages} to: a) train the linear projector and b) retrieve images during inference. Since the work focuses on human generation, we run a face detector~\cite{deng2019retinaface} and only keep the images with human faces. Since the datasets contain low quality images \eg, blurry scenes, we run an aesthetic scorer~\cite{christophschuhmann} to filter out images with low scores. We combine the MSCOCO and OpenImages datasets, splitting into non-overlapping training and retrieval sets, consisting of 173,289 and 330,777 images respectively. For retrieval, we index the image embeddings using CLIP VIT-L/14~\cite{Ilharco_OpenCLIP_2021}. Since these embeddings are pre-computed and stored, \gls{FRAG} avoids using CLIP image encoder during inference.

\subsection{Comparison Methods}
\label{sec:comparison_methods}
We compare \gls{FRAG} against these methods:

\begin{itemize}[leftmargin=*]
    \item \textbf{\textit{\gls{SD}}} is the baseline method used without applying any debiasing technique.
    \item \textbf{Ethical Interventions (\textit{\gls{Interven}})} 
    attempts to improve diversity by augmenting the original prompts with ethical phrases, \eg, \textit{Photo of a doctor if all individuals can be a doctor irrespective of their age, gender and skin tone}.
    
    \item \textbf{Fair Diffusion (\textit{\gls{FairDiff}})} applies semantic-guidance~\cite{brack2023sega} to steer the model towards a specific intersectional group. The groups are sampled with a uniform prior.
    
    \item \textbf{Text Augmentation (\textit{\gls{TextAug}})} creates multiple variants of the prompt by explicitly mentioning the demographic groups, \eg, \textit{Photo of a doctor. This person is 55-year old, dark-skinned, female}. Surprisingly, despite its simplicity, past studies do not study this method~\cite{bansal2022well,friedrich2023fair}. We find \gls{TextAug} to be a strong baseline in our experiments.
    
    \item \textbf{Baseline RAG (\textit{\gls{BaseRAG}})} is an ablated version of \gls{FRAG} that removes the fairness interventions, \ie, does not use debiased query, balanced sampling or text instruction. It still relies on the linear module for conditioning.
    
\end{itemize}

\gls{FairDiff} and \gls{TextAug} require explicit specifications of the demographic groups. For this, we use the template: \textit{This person is $<$age$>$-year old, 
$<$skin tone$>$, $<$gender$>$}. Age is the mid-point of the corresponding group \eg, \textit{25-year-old} for the group: \textit{20–29 years old}; skin tone is specified as: \textit{light-skinned}, \textit{medium skin colored} or \textit{dark-skinned} and gender is either \textit{male} or \textit{female}. Note that \gls{FRAG} avoids such explicit attribute specification, relying instead on the text instruction for implicit attribute transfer.

\begin{figure*}[t]
\footnotesize
 \centering
    \includegraphics[width=0.88\textwidth]{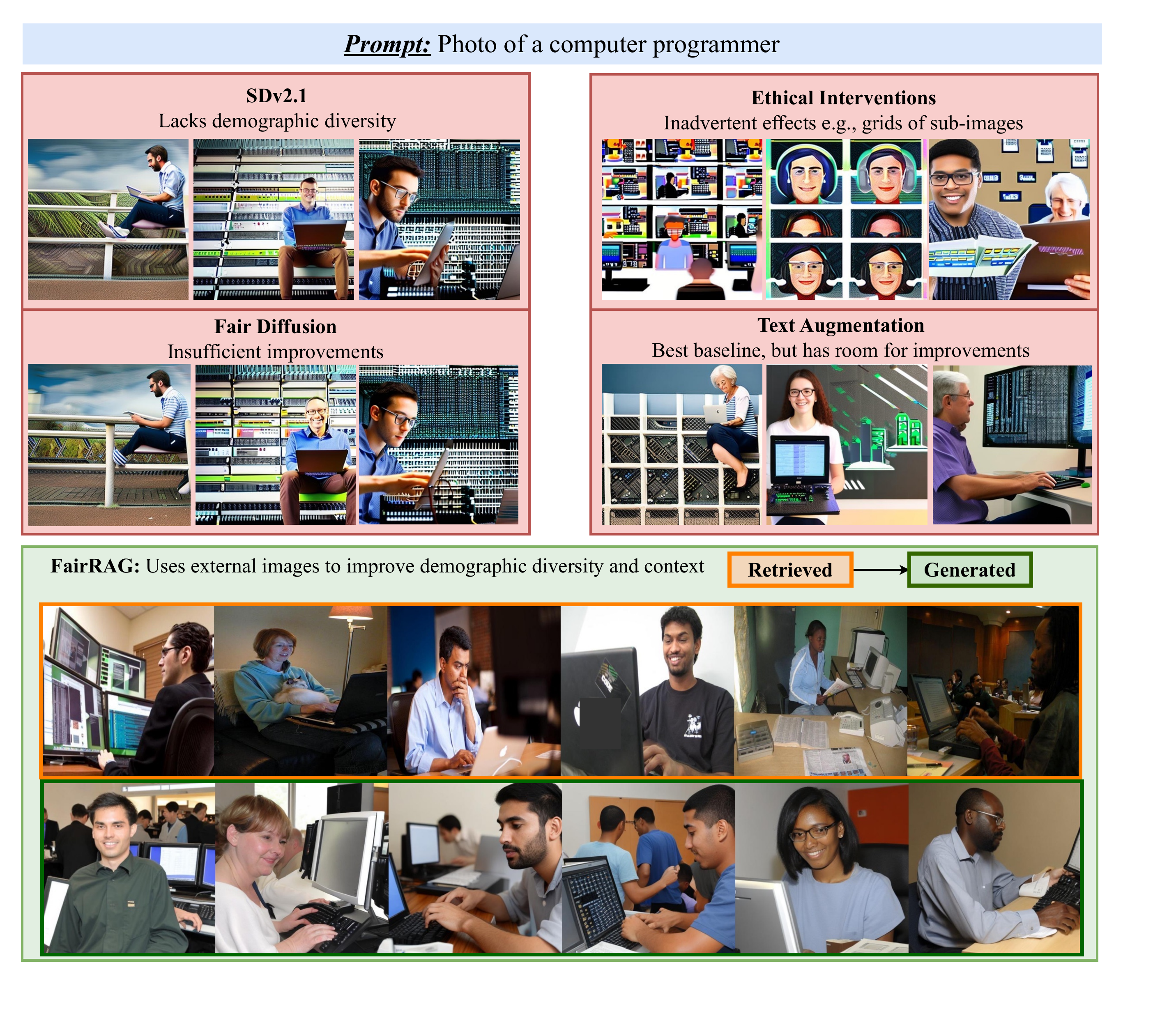}
  \caption{Example outputs from different methods for the text prompt \textit{Photo of a computer programmer}. Baseline methods, barring Text Augmentation, fail to produce images with high demographic diversity. \gls{FRAG} improves demographic diversity with the help of external visual references. Apart from that, it also improves alignment and fidelity.}
\label{fig:qual_comp}
\vspace{-2mm}
\end{figure*}

\begin{table}[t]
\centering
\caption{Breakdown of diversity scores for individual and intersectional (intersec.) groups, showing how leveraging external images can help improve the metrics.}
\label{tab:diversity}
\begin{tabular}{lcccc} \hline
\textbf{Method}      & \textbf{\begin{tabular}[c]{@{}c@{}}Age \end{tabular}} & \textbf{\begin{tabular}[c]{@{}c@{}}Gender\end{tabular}} & \textbf{\begin{tabular}[c]{@{}c@{}}Skin\\ Tone\end{tabular}} & \textbf{\begin{tabular}[c]{@{}c@{}}Intersec.\end{tabular}} \\ \hline
\gls{SD} & 0.220 & 0.273 & 0.224 & 0.188 \\
\gls{Interven} & \underline{0.439} & 0.451 & 0.362 & 0.333 \\
\gls{FairDiff} & 0.225 & 0.371 & 0.223 & 0.196 \\
\gls{TextAug} & 0.426 & \underline{0.766} & 0.334 & 0.341 \\
\gls{BaseRAG} & 0.417 & 0.582 & \textbf{0.439} & \underline{0.374} \\
\gls{FRAG} & \textbf{0.559} & \textbf{0.800} & \underline{0.416} & \textbf{0.438} \\
 \hline
\end{tabular}
\end{table}

\subsection{Results}

In this section, we discuss the overall quantitative and qualitative results. Table~\ref{tab:overall_results} summarizes the intersectional diversity, alignment and fidelity metrics alongside the absolute gains over \gls{SD}. \gls{FRAG} outperforms other methods in terms of the diversity and alignment scores and is close to \gls{BaseRAG} in terms of image fidelity, showing the benefits of the proposed setup. As shown in Fig.~\ref{fig:qual_comp}, \gls{FRAG} effectively leverages human-specific attributes from the reference images to condition the generated images, resulting in enhanced demographic diversity. As presented in Table~\ref{tab:diversity}, it is able to improve diversity metrics for all three attributes: age, gender and skin tone. In terms of the baselines, we find that \gls{BaseRAG} is also able to improve diversity, alignment and fidelity scores, however, the additional fairness interventions used in \gls{FRAG}, \ie, \textit{query debiasing}, \textit{balanced sampling} and the \textit{transfer instruction} help boost the diversity scores further. In terms of non-RAG baselines, we find \gls{TextAug} to be the most effective, obtaining improvements in all three metrics over other non-RAG methods. However, qualitatively, it produces synthetic, unrealistic contexts, an issue observed for other baselines as well. \gls{FRAG} on the other hand generates more realistic images due to the conditioning from real images. Next, \gls{Interven} and \gls{FairDiff} also improve the diversity scores to some extent, but are well below \gls{FRAG}. As shown in Fig.~\ref{fig:qual_comp}, extra text intervention used in \gls{Interven} results in grids of smaller sub-images, which is an undesired side-effect. \gls{FRAG} also uses a text instruction, but this does not lead to such inadvertent consequences. Therefore compared to the baseline methods, \gls{FRAG} stands out as more effective.

\begin{table}[t]
\centering
\caption{Quantitative results from all the comparison methods, highlighting the \textbf{best} and the \underline{second-best} scores. We also show the \textcolor{green}{improvement} and \textcolor{red}{deterioration} in terms of absolute difference from SDv2.1. Compared to other baselines, \gls{FRAG} shows improvements in diversity and  alignment, and rivals \gls{BaseRAG} in terms of the fidelity score.}
\label{tab:overall_results}
\setlength\tabcolsep{3 pt}
\begin{tabular}{llll} \hline
\textbf{Method}      &
\textbf{\begin{tabular}[c]{@{}c@{}}Diversity  ($\uparrow$)\end{tabular}} &
\textbf{\begin{tabular}[c]{@{}c@{}}CLIP ($\uparrow$)\end{tabular}} & \textbf{\begin{tabular}[c]{@{}c@{}}FID ($\downarrow$)\end{tabular}} \\ \hline
\gls{SD} & 0.188 & 0.142 & 85.3 \\
\gls{Interven} & 0.333 (\textcolor{green} {+.145}) & 0.132 (\textcolor{red} {-.011}) & 93.9 (\textcolor{red} {+08.6}) \\
\gls{FairDiff} & 0.196 (\textcolor{green} {+.008}) & 0.142 (\textcolor{red} {-.000}) & 87.8 (\textcolor{red} {+02.5}) \\
\gls{TextAug} & 0.341 (\textcolor{green} {+.153}) & 0.\underline{144} (\textcolor{green} {+.002}) & 74.1 (\textcolor{green} {-11.2}) \\
\gls{BaseRAG} & \underline{0.374} (\textcolor{green} {+.186}) & \textbf{0.146} (\textcolor{green} {+.003}) & \textbf{49.4} (\textcolor{green} {-35.9}) \\
\gls{FRAG} & \textbf{0.438} (\textcolor{green} {+.250}) & \textbf{0.146} (\textcolor{green} {+.003}) & \underline{51.8} (\textcolor{green} {-33.5}) \\
\hline                      \end{tabular}
\end{table}

\begin{figure*}[t]
\footnotesize
 \centering
    \includegraphics[width=0.85\textwidth]{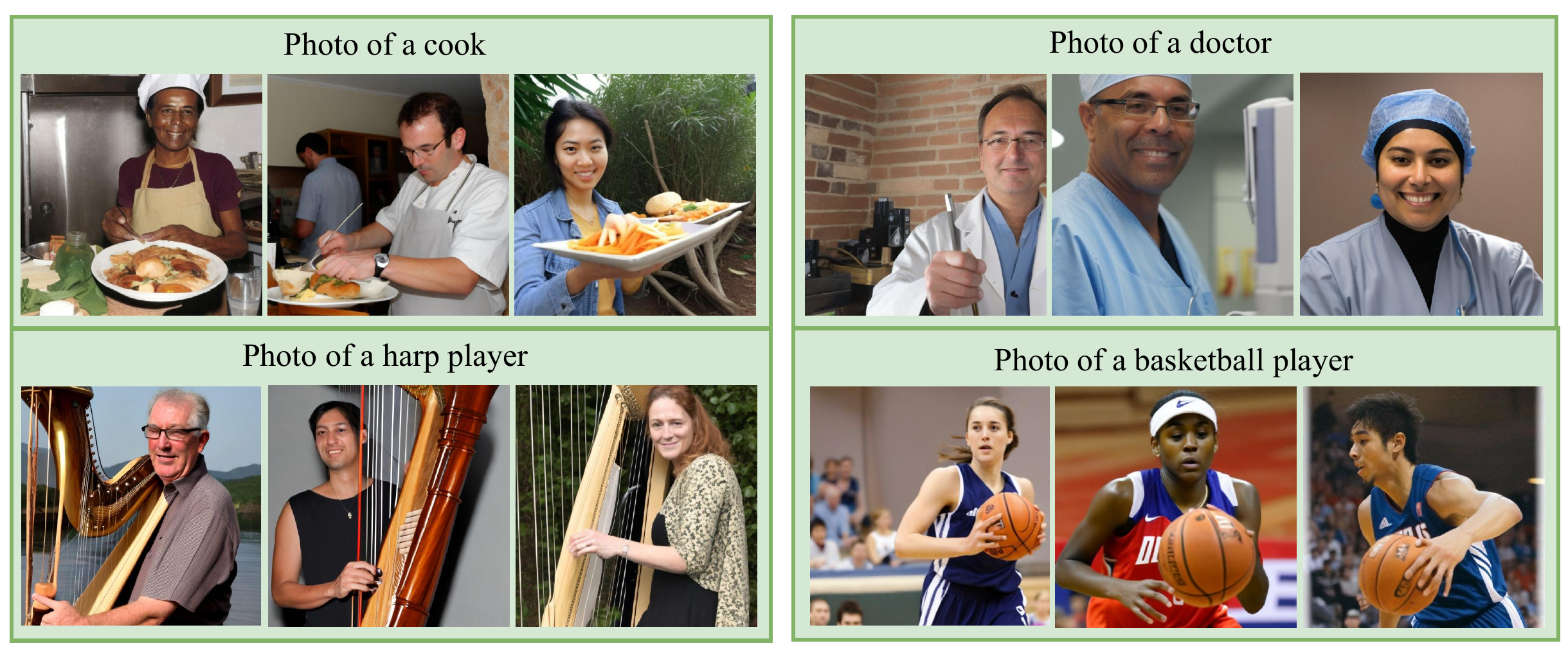}
  \caption{\gls{FRAG} improves demographic diversity for different categories of professions.}
\label{fig:qual_FRAG}
\end{figure*}

\noindent \textbf{Diversity for different prompt categories.} \quad We present example outputs for different categories in Fig.~\ref{fig:qual_FRAG} and present the intersectional diversity values for each of the 8 categories in Table~\ref{tab:by_concept_type}. The improved diversity metrics shows that \gls{FRAG} generalizes to different professions.

\noindent \textbf{Minimal Increase in Latency.} \quad \gls{FRAG} involves: a) text-to-image retrieval with debiased query, b) balanced sampling, c) visual reference projection to obtain the bimodal prompt, and d) conditional image generation. The first three steps are specific to \gls{FRAG}, but add minimal computational overhead compared to~\gls{SD_nocite}. On a single NVIDIA A10G Tensor Core GPU, \gls{SD_nocite} and \gls{FRAG} require 2.8 secs and 3 secs respectively, to generate a single image with 20 denoising steps. We re-iterate that \gls{FRAG} is also more efficient than prior methods that use test-time tuning~\cite{gal2022image} or heavier conditioning modules~\cite{zhang2023adding}.

\begin{figure}[t]
\footnotesize
 \centering
    \includegraphics[width=0.48\textwidth]{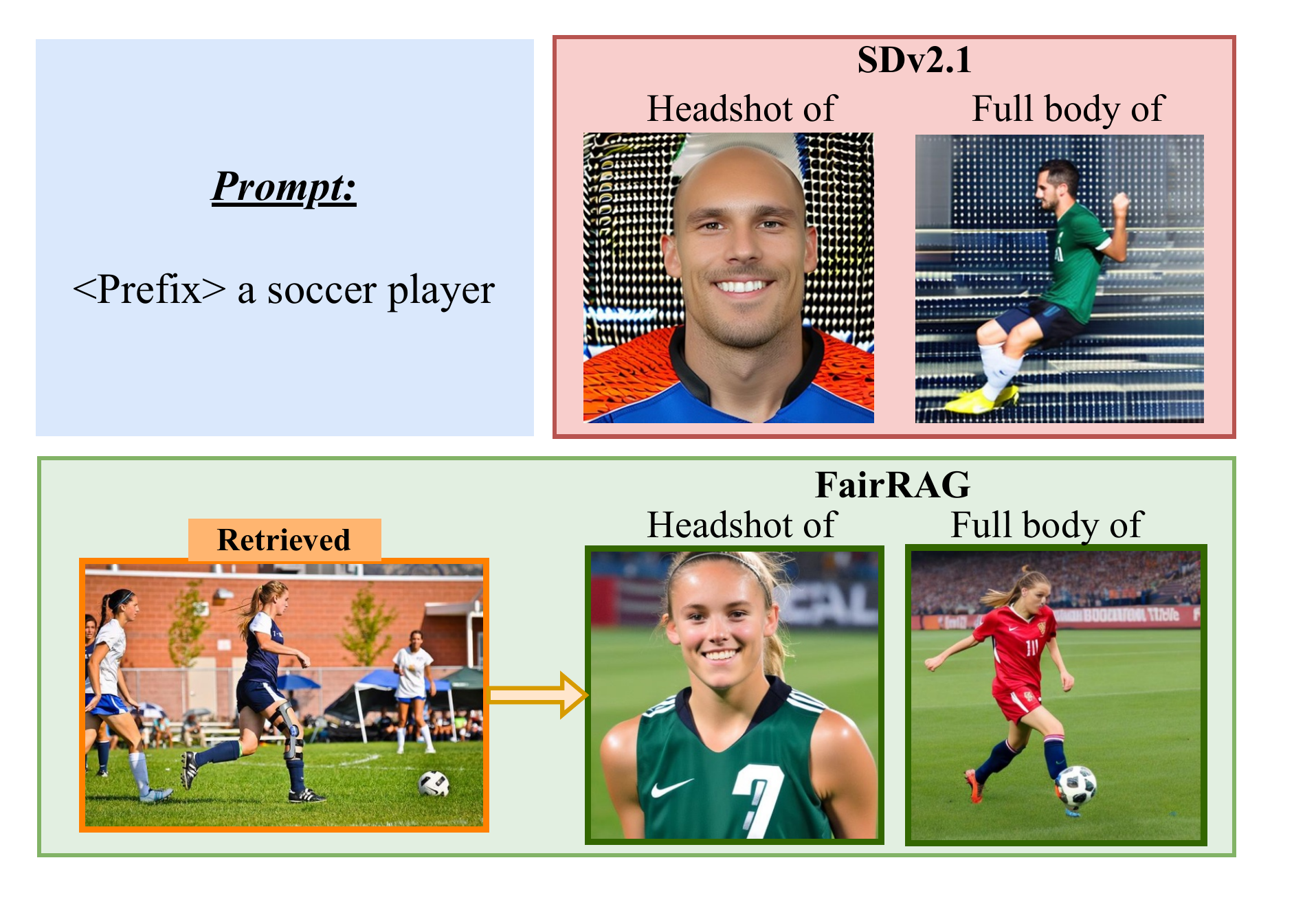}
  \caption{While most past works focus on close-up views of faces, we find \gls{FRAG} can transfer attributes when asked to generate full body images as well.}
\label{fig:face_size}
\end{figure}

\noindent \textbf{Face and Body Size.} \quad Most past studies focus on close-up views of faces neglecting analysis of images with the human body taking a larger portion of the image~\cite{friedrich2023fair,zhang2023iti}. To test if \gls{FRAG} works well for both the cases, we employ two different prompt prefixes: \textit{Headshot of} and \textit{Full body of}, controlling the face/body size. As shown in Fig.~\ref{fig:face_size}, we find that \gls{FRAG} is able to transfer the demographic attributes in both the cases. It also generates contexts that are more realistic than the SDv2.1 baseline, which is especially evident for full body images.

\noindent \textbf{Ablation Study.} \quad In Table~\ref{tab:ablations}, we present ablated variants of \gls{FRAG} to investigate the effects of different components. Retrieval-time interventions: \textit{debiased query} and \textit{balanced sampling} and generation-time intervention: \textit{text instruction}, all contribute positively to the intersectional diversity score, thereby validating our decisions to incorporate these mechanisms. All three mechanisms enhance the age and diversity scores. For skin tone diversity, we do not find additional benefit from text instruction, but debiased query and balanced sampling contribute positively to skin tone diversity as well. We also present the diversity scores for the real distribution, \ie, the retrieved images. \gls{FRAG} still has a room for improvement, which can potentially be achieved by improving the transfer of attributes.

\begin{table*}[t]
\centering
\caption{Intersectional diversity metrics on the eight concept types used in our evaluation set. \gls{FRAG} shows improvements in each concept type, showcasing the generality of the approach.}
\label{tab:by_concept_type}
\begin{tabular}{lcccccccc} \hline
& \textbf{\begin{tabular}[c]{@{}c@{}}Artists\end{tabular}} & \textbf{\begin{tabular}[c]{@{}c@{}}F\&B \\ Workers\end{tabular}} & \textbf{\begin{tabular}[c]{@{}c@{}}Musicians\end{tabular}} & \textbf{\begin{tabular}[c]{@{}c@{}}Security \\ Personnel\end{tabular}} & \textbf{\begin{tabular}[c]{@{}c@{}}Sports \\ Players \end{tabular}} & \textbf{\begin{tabular}[c]{@{}c@{}}STEM \\ Profes.\end{tabular}} & \textbf{\begin{tabular}[c]{@{}c@{}} Workers\end{tabular}} & \textbf{\begin{tabular}[c]{@{}c@{}}Others\end{tabular}} \\ \hline
\gls{SD} & 0.261 & 0.237 & 0.168 & 0.137 & 0.175 & 0.199 & 0.197 & 0.175 \\
\gls{Interven} & 0.385 & 0.284 & 0.282 & 0.314 & 0.299 & 0.359 & 0.282 & 0.370 \\
\gls{FairDiff} & 0.259 & 0.273 & 0.164 & 0.133 & 0.161 & 0.240 & 0.210 & 0.177 \\
\gls{TextAug} & 0.391 & 0.269 & 0.322 & 0.348 & 0.314 & 0.342 & 0.349 & 0.357 \\
\gls{BaseRAG} & \underline{0.401} & \textbf{0.428} & \underline{0.404} & \underline{0.394} & \underline{0.336} & \underline{0.357} & \underline{0.362} & \underline{0.402} \\
\gls{FRAG} & \textbf{0.436} & \underline{0.413} & \textbf{0.440} & \textbf{0.458} & \textbf{0.416} & \textbf{0.432} & \textbf{0.419} & \textbf{0.454} \\
\hline                      \end{tabular}
\end{table*}

\section{Limitations and Future Directions}
In this section, we discuss some limitations and layout potential future directions for further improvement. To begin with, \gls{FRAG} uses one-to-one image mapping \ie uses single reference image for each generated image. An alternative would be to use multiple images to summarize the concepts to be transferred to enhance the conditioning process. Multiple references could also help in cases where single retrieval does not encompass all of the concepts mentioned in the prompt, by aggregating different concepts from different images. Second, despite conditioning on real images, the samples generated by \gls{FRAG} can contain disfigurements especially in small faces, limbs and fingers. We hypothesize that fixing this issue requires a better way to incorporate knowledge on human anatomy within the models, which likely entails re-training or tuning the backbone. We discuss this issue in greater detail in~\ref{sec:disfigurements}.

There are other considerations before a framework such as \gls{FRAG} can be deployed in practice. For one, \gls{FRAG} is limited to human image generation and thus cannot tackle non-human prompts. A more general RAG framework could utilize references from a broader range of categories. However, even with a larger data source, a practical framework should also be capable of tackling cases in which the references lack consistency with the user's prompt. A worthwhile future direction to tackle this issue is to devise conditioning mechanisms that avoid transferring concepts in references that conflict with the prompt. Furthermore, while the scope of this study is limited to generating humans, we note that the proposed conditioning mechanism and the fairness interventions can be extended to and employed in other domains, making these mechanisms more general with broader applicability.

\begin{table}[t]
\centering

\caption{Ablation studies showing how debiased query (debiased q), balanced sampling (bal. sampl.) and text instruction (text instr.) help boost the diversity scores. We also present the metrics for retrieved images, \ie, the real distribution, showcasing room for further improvement.}
\label{tab:ablations}
\begin{tabular}{lcccc} \hline
\textbf{Method}      & \textbf{\begin{tabular}[c]{@{}c@{}}Age \end{tabular}} & \textbf{\begin{tabular}[c]{@{}c@{}}Gender\end{tabular}} & \textbf{\begin{tabular}[c]{@{}c@{}}Skin\\ Tone\end{tabular}} & \textbf{\begin{tabular}[c]{@{}c@{}}Intersec.\end{tabular}} \\ \hline
\gls{SD} & 0.220 & 0.273 & 0.224 & 0.188 \\
\gls{TextAug} & 0.426 & 0.766 & 0.334 & 0.341 \\ 
\gls{BaseRAG} & 0.440 & 0.562 & \underline{0.437} & 0.386 \\
\hline
\multicolumn{5}{c}{\textit{Ablated variants of \gls{FRAG}}} \\
No Debiased Q & 0.525 & 0.764 & 0.411 & 0.414 \\
No Bal. Sampl. & 0.538 & 0.734 & 0.392 & 0.420 \\
No Text Instr. & 0.481 & 0.771 & 0.416 & 0.407 \\
\gls{FRAG} & \textbf{0.559} & \underline{0.800} & 0.416 & \underline{0.438} \\ \hline
Retrievals & \underline{0.546} & \textbf{0.902} & \textbf{0.526} & \textbf{0.478} \\ \hline
\end{tabular}
\end{table}

\section{Conclusion}
In this work, we developed the \gls{FRAG} framework to condition pre-trained generative models on external images to improve demographic diversity. We showed that a lightweight, linear layer can be trained to project visual references for conditioning the backbone and post-hoc debiasing methods can enhance fairness in generation. These mechanisms add minimal overhead during inference, yet, help \gls{FRAG} surpass prior methods in terms of diversity, alignment and fidelity.

\clearpage
{
    \small
    \bibliographystyle{ieeenat_fullname}
    \bibliography{main}
}

\clearpage
\section*{Appendix}

\appendix

\section{Additional Results}

\subsection{Qualitative Analysis} 
\begin{figure*}[t]
  \strut\newline
  \centering
  
   \includegraphics[width=0.95\textwidth]{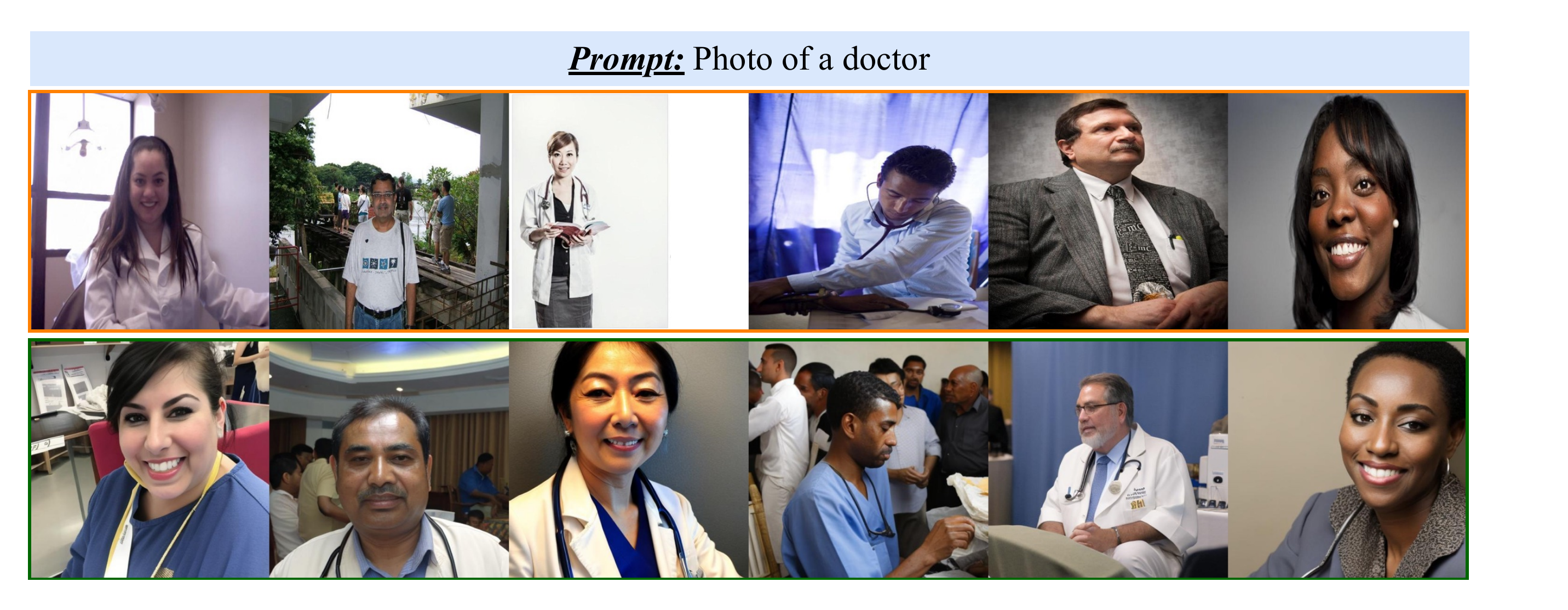}
     \includegraphics[width=0.95\textwidth]{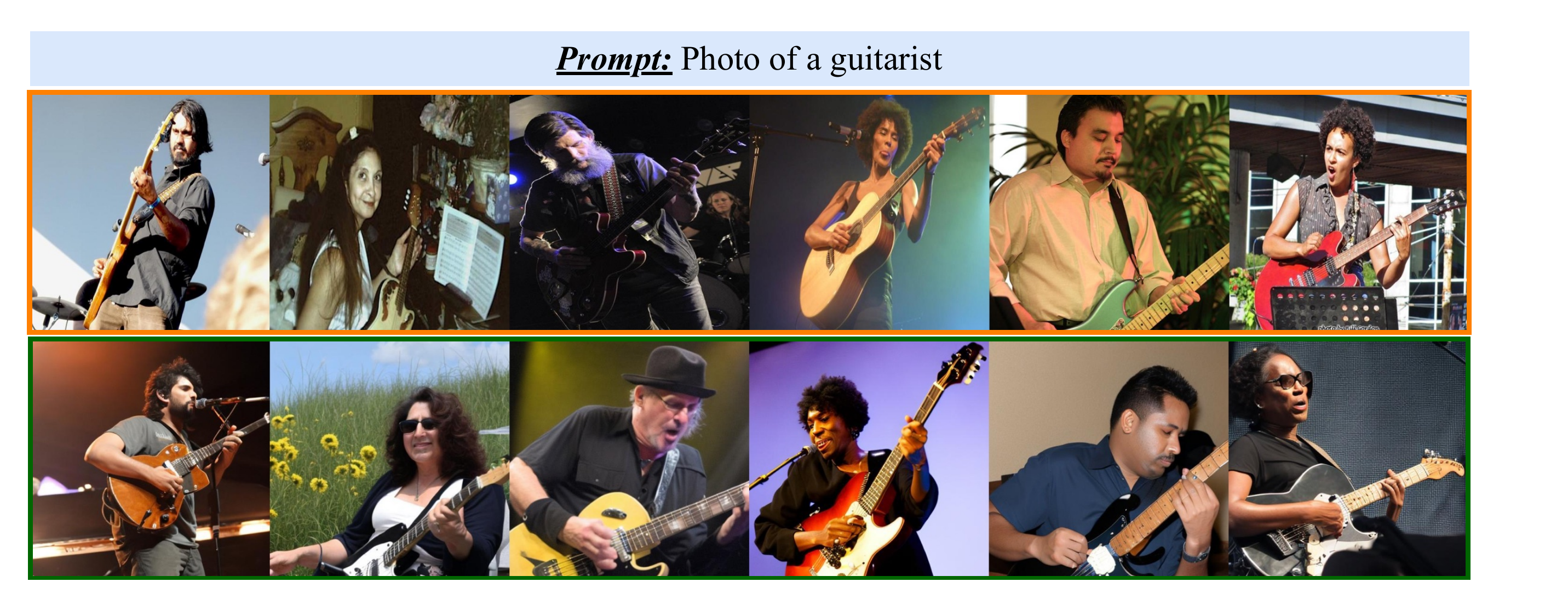}
     \includegraphics[width=0.95\textwidth]{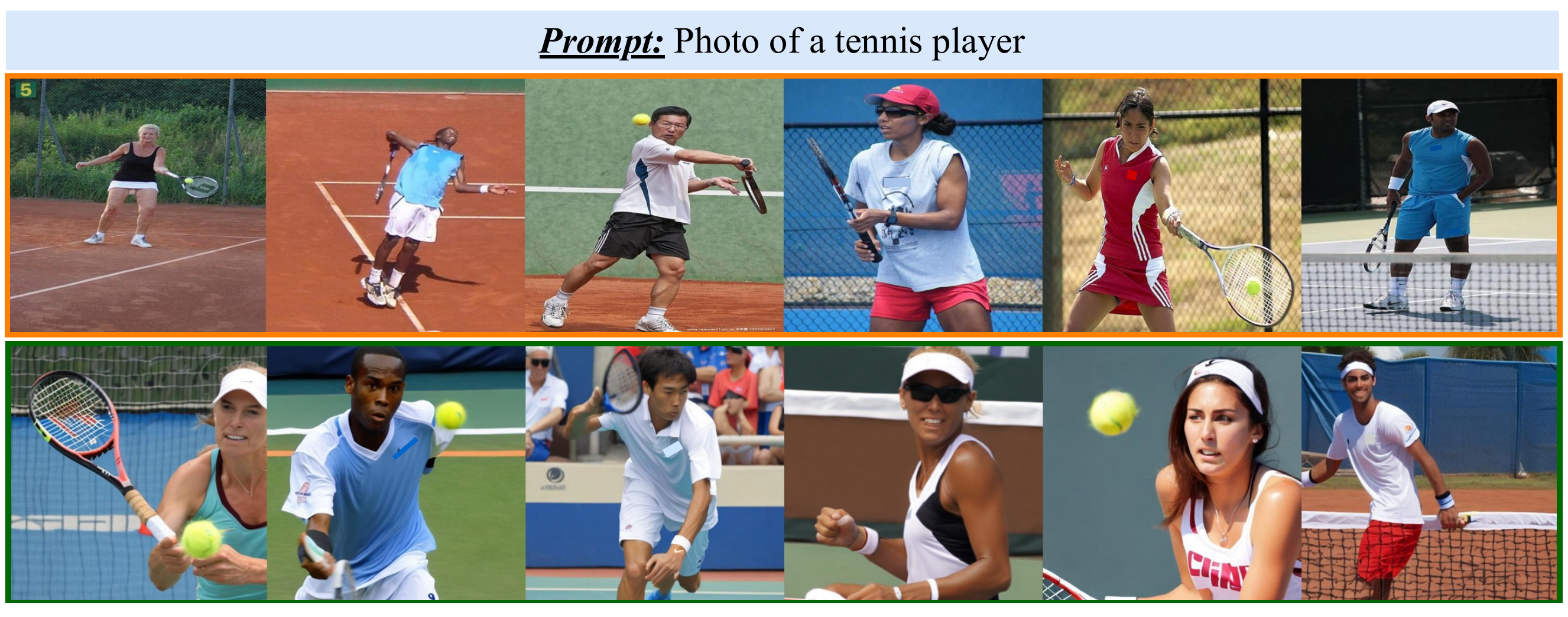}
  \captionof{figure}{Example outputs illustrating how  \gls{FRAG} uses the reference images (highlighted in orange) to improve diversity of the generated images (highlighted in green). }\label{fig:qual_appendix}
\end{figure*}

We present more qualitative examples including the reference (highlighted in orange) and generated (highlighted in green) images from \gls*{FRAG} in Fig.~\ref{fig:qual_appendix}. We observe that \gls*{FRAG} is able to utilize the reference images to improve demographic diversity.

\subsection{Quantitative Analysis} 
We present comprehensive results in Table~\ref{tab:ablations_full} with all the metrics for all the non-RAG baselines alongside the different variants of \gls{FRAG}. We present results for both retrieved and generated images. First, we observe that the intersectional diversity scores improve for both real and generated distributions with \textit{debiased query}, \textit{balanced sampling} and \textit{text instruction}. We also observe some trade-offs between the diversity and alignment/fidelity metrics. CLIP score increases when debiased query is not used and FID value improves when text instruction is not used, while both showcase improvements in the diversity score. This leaves a room for improvement in both alignment and fidelity with the additional mechanisms.

\subsection{Disfigurements}
\label{sec:disfigurements}
\begin{figure}[t]
\footnotesize
 \centering
    \includegraphics[width=0.46\textwidth]{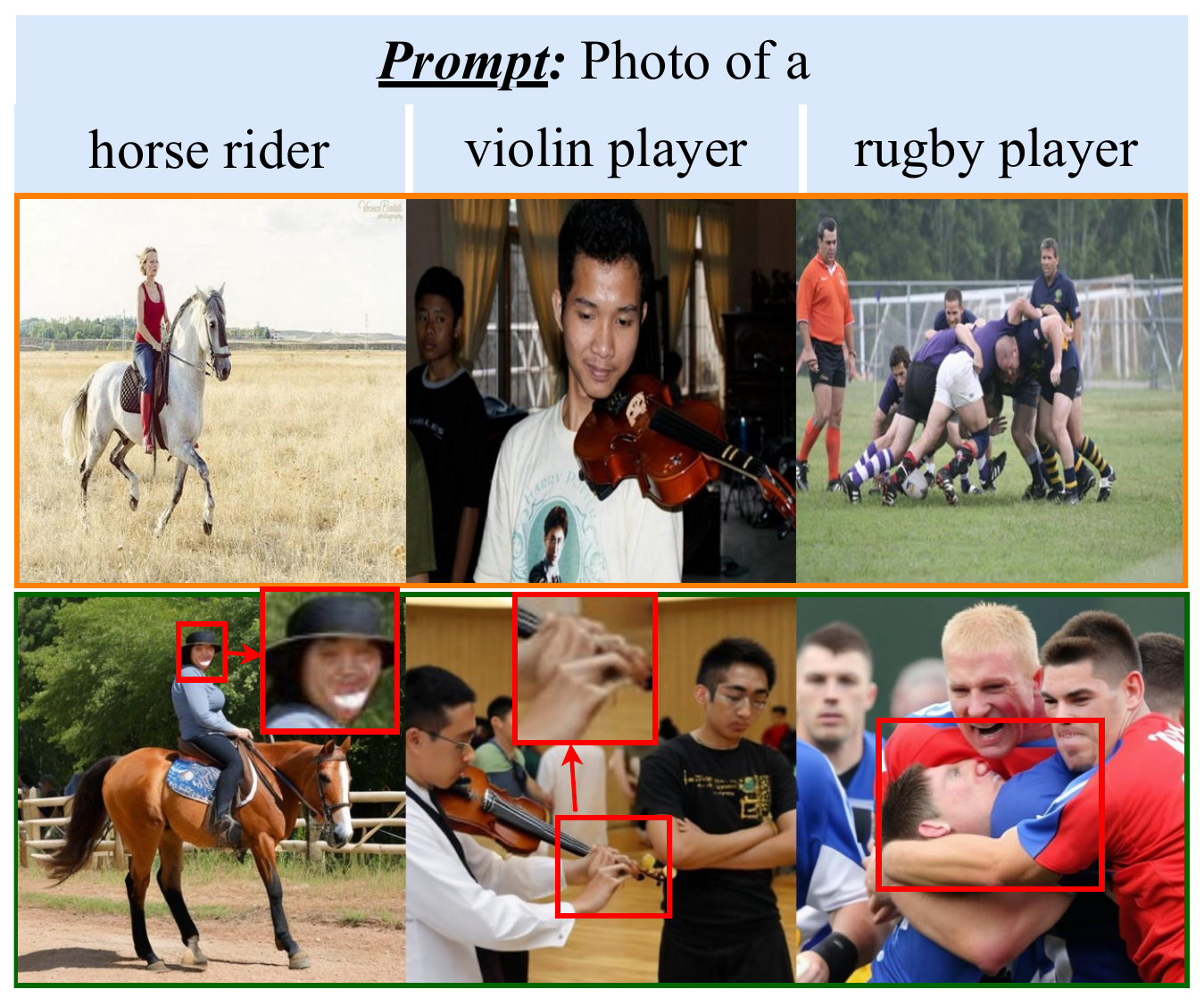}
  \caption{Despite conditioning on real images, the outputs from \gls{FRAG} can still contain disfigurements as depicted within the red boxes. Fixing this issue likely requires improved mechanisms to incorporate the knowledge on human anatomy in the models.}
\label{fig:disfigured}
\end{figure}

As shown in Fig.~\ref{fig:disfigured}, the generated images can contain disfigurements for small faces, limbs and fingers. We address this issue to a limited extent by using a negative prompt: \textit{bad, disfigured, cropped, bad anatomy, poorly drawn hands, poorly drawn fingers} for all the methods. Simply conditioning frozen backbone on real images does not solve this issue. We hypothesize further improvements require incorporation of the knowledge on human anatomy within the models, which likely entails re-training or tuning the backbone. We leave this for future research efforts.

\subsection{Varying number of candidates ($N$)} 
In Table~\ref{tab:topN}, we analyze the effects of the initial number of candidates $N$ used to gather the subset of $K$ references. Diversity score increases from $N$ = 100 to $N$ = 750, and saturates after that. We do not observe clear trends for CLIP and FID scores. For all the experiments, we set $N = 250$, using results from preliminary experiments without tuning $N$ on the test set. We set $K = 20$ to compute all the metrics.

\section{Evaluation Set}
\label{appendix:eval_set}
The evaluation set consists of 80 prompts that exhibit bias with respect to different demographic groups. They are classified into 8 categories, including:
\begin{itemize}
    \item \textbf{6 Artists}: craftsperson, dancer, makeup artist, painter, puppeteer, sculptor
    \item \textbf{6 Food and Beverage Workers}: bartender, butcher, chef, cook, fast-food worker, waiter
    \item \textbf{9 Musicians}: disk jockey, drummer, flutist, guitarist, harp player, keyboard player,  singer, trumpeter, violin player
    \item \textbf{6 Security Personnels}: firefighter, guard, lifeguard, police officer, prison officer, soldier
    \item \textbf{9 Sports Players}: baseball player, basketball player, gymnast, horse rider, rugby player, runner, skateboarder, soccer player, tennis player
    \item \textbf{12 STEM Professionals}: architect, astronaut, computer programmer, dentist, doctor, electrician, engineer, mechanic, nurse, pilot, scientist, surgeon
    \item \textbf{7 Workers}: carpenter, farmer, gardener, housekeeper, janitor, laborer, person washing dishes
    \item \textbf{25 Others}: backpacker, cashier, CEO, cheerleader, climber, flight attendant, hairdresser,  judge, lawyer, lecturer, motorcyclist, patient, politician, public speaker, referee, reporter,  retailer, salesperson, sailor, seller,  social worker, solicitor, student, tailor, teacher
\end{itemize}

\section{Implementation Details}
We train the linear encoder: $\mathcal{H}$ for 50K iterations using the AdamW optimizer~\cite{loshchilov2017decoupled} ($\beta_1 = 0.9, \beta_2=0.999$), with a learning rate of 1e-3 and a weight decay of 0.01. We use balanced sampling during training with a uniform prior over each intersectional group (age, gender and skin tone). During training, we clip the gradients if the norm is greater than 1.0. To generate images during inference, we use the DDIM noise scheduler~\cite{song2020denoising}, with 20 de-noising steps conditioned on the text prompt, textual instruction and the projected visual reference.

\begin{table*}[t]
\caption{Presenting diversity, alignment and fidelity metrics for all the baselines and ablated versions of \gls{FRAG}. We present results for both retrieved and generated images for \gls{FRAG}. }
\centering
\label{tab:ablations_full}
\begin{tabular}{lcccccc}  \hline
 & \multicolumn{4}{c}{\textbf{Diversity}} & \multicolumn{1}{c}{\multirow{2}{*}{\textbf{CLIP}}} & \multicolumn{1}{c}{\multirow{2}{*}{\textbf{FID}}} \\
 & \multicolumn{1}{c}{\textbf{Age}} & \multicolumn{1}{c}{\textbf{Gender}} & \multicolumn{1}{c}{\textbf{Skin Tone}} & \multicolumn{1}{c}{\textbf{Intersec.}} & \multicolumn{1}{c}{} & \multicolumn{1}{c}{} \\  \hline
\gls{SD} & 0.220 & 0.273 & 0.224 & 0.188 & 0.142 & 85.3 \\
\gls{Interven} & 0.439 & 0.451 & 0.362 & 0.333 & 0.132 & 93.9 \\
\gls{FairDiff} & 0.225 & 0.371 & 0.223 & 0.196 & 0.142 & 87.8 \\
\gls{TextAug} & 0.426 & 0.766 & 0.334 & 0.341 & 0.144 & 74.1 \\ \hline
\multicolumn{7}{c}{\textit{\textbf{Ablated variants of FairRAG}}} \\ \hline
\multicolumn{7}{c}{\textit{BaseRAG}} \\
Retrieved & 0.475 & 0.622 & 0.558 & 0.447 & 0.167 & 33.1 \\
Generated & 0.440 & 0.562 & 0.437 & 0.386 & 0.146 & 49.4 \\ \hline
\multicolumn{7}{c}{\textit{Without Debiased Query}} \\
Retrieved & 0.477 & 0.867 & 0.530 & 0.460 & 0.166 & 31.9 \\
Generated & 0.525 & 0.764 & 0.411 & 0.414 & 0.150 & 50.5 \\  \hline
\multicolumn{7}{c}{\textit{Without Balanced Sampling}} \\
 Retrieved & 0.528 & 0.741 & 0.522 & 0.458 & 0.159 & 30.0 \\ 
Generated & 0.538 & 0.734 & 0.392 & 0.420 & 0.146 & 53.0 \\  \hline
\multicolumn{7}{c}{\textit{Without Text Instruction}} \\
Retrieved & 0.544 & 0.902 & 0.526 & 0.478 & 0.158 & 26.5 \\ 
Generated & 0.481 & 0.771 & 0.416 & 0.407 & 0.145 & 48.9 \\  \hline
\multicolumn{7}{c}{\textit{FairRAG}} \\
Retrieved & 0.544 & 0.902 & 0.526 & 0.478 & 0.158 & 26.5 \\
Generated & 0.559 & 0.800 & 0.416 & 0.438 & 0.146 & 51.8 \\ \hline
\end{tabular}
\end{table*}

\begin{table*}[t]
\caption{Diversity, image-text alignment and image fidelity metrics for different values of $N$ used for retrieval.}
\centering
\label{tab:topN}
\begin{tabular}{lcccccc}  \hline
 & \multicolumn{4}{c}{\textbf{Diversity}} & \multicolumn{1}{c}{\multirow{2}{*}{\textbf{CLIP}}} & \multicolumn{1}{c}{\multirow{2}{*}{\textbf{FID}}} \\
 & \multicolumn{1}{c}{\textbf{Age}} & \multicolumn{1}{c}{\textbf{Gender}} & \multicolumn{1}{c}{\textbf{Skin Tone}} & \multicolumn{1}{c}{\textbf{Intersec.}} & \multicolumn{1}{c}{} & \multicolumn{1}{c}{} \\  \hline
\gls{SD} & 0.220 & 0.273 & 0.224 & 0.188 & 0.142 & 85.3 \\
\gls{Interven} & 0.439 & 0.451 & 0.362 & 0.333 & 0.132 & 93.9 \\
\gls{FairDiff} & 0.225 & 0.371 & 0.223 & 0.196 & 0.142 & 87.8 \\
\gls{TextAug} & 0.426 & 0.766 & 0.334 & 0.341 & 0.144 & 74.1 \\ \hline
\multicolumn{7}{c}{\textit{\textbf{Top-N}}} \\ \hline
N=100 & 0.547 & 0.785 & 0.409 & 0.433 & 0.145 & 52.7 \\
N=250 & 0.559 & 0.800 & 0.416 & 0.438 & 0.146 & 51.8 \\
N=500 & 0.580 & 0.816 & 0.407 & 0.443 & 0.145 & 51.5 \\
N=750 & 0.586 & 0.824 & 0.415 & 0.447 & 0.144 & 52.2 \\
N=1000 & 0.572 & 0.850 & 0.418 & 0.445 & 0.146 & 52.8 \\ \hline
\end{tabular}
\end{table*}

\end{document}